\documentclass[10pt,twocolumn,letterpaper,usenames]{article}

\usepackage[usenames,dvipsnames]{color}
\usepackage{cvpr}
\usepackage{times}
\usepackage{epsfig}
\usepackage{graphicx}
\usepackage{amsmath}
\usepackage{amssymb}

\usepackage{booktabs}
\usepackage{wasysym}
\usepackage[caption=false]{subfig}

\makeatletter
\newcommand{\tmpnoeso}[1]{\let\@temp\ESO@HookI \global\let\ESO@HookI\@empty #1 \global\let\ESO@HookI\@temp}
\makeatother

\usepackage[printonlyused,nolist,nohyperlinks]{acronym}
\usepackage{xcolor}

\usepackage{pgfplots}
\pgfplotsset{compat=newest}
\pgfplotsset{plot coordinates/math parser=false}
\usepgfplotslibrary{external}
\usetikzlibrary{backgrounds}

\usepackage[breaklinks=true,bookmarks=false]{hyperref}

\cvprfinalcopy 

\def\Kinect{Kinect}

\def\cvprPaperID{1981} 

\begin{document}

\title{Learning Descriptors for Object Recognition and 3D Pose Estimation}

\author{Paul Wohlhart and Vincent Lepetit\\
Institute for Computer Vision and Graphics, 
Graz University of Technology, Austria\\
{\tt\small \{wohlhart,lepetit\}@icg.tugraz.at}
}

\maketitle
\thispagestyle{empty}

\begin{abstract}

Detecting  poorly  textured  objects  and  estimating  their  3D  pose
reliably is still  a very challenging problem.  We  introduce a simple
but powerful approach  to computing descriptors for  object views that
efficiently capture both the object identity and 3D pose.  By contrast
with previous manifold-based approaches, we  can rely on the Euclidean
distance to evaluate the similarity between descriptors, and therefore
use scalable Nearest  Neighbor search methods to  efficiently handle a
large number  of objects  under a  large range  of poses.   To achieve
this,  we  train  a  Convolutional Neural  Network  to  compute  these
descriptors   by  enforcing   simple   similarity  and   dissimilarity
constraints  between the  descriptors.  We  show that  our constraints
nicely untangle the images from different objects and different views
into clusters that are not  only well-separated but also structured as
the  corresponding  sets  of  poses: The  Euclidean  distance  between
descriptors is large when the  descriptors are from different objects,
and  directly related  to  the  distance between  the  poses when  the
descriptors  are from  the  same object.   These important  properties
allow us  to outperform state-of-the-art object  views representations
on challenging RGB and RGB-D data.

\end{abstract}

\section{Introduction}

Impressive results  have been  achieved in  3D pose  estimation of  objects from
images  during the  last decade.   However, current  approaches cannot  scale to
large-scale  problems  because  they  rely  on one  classifier  per  object,  or
multi-class classifiers such as Random  Forests, whose complexity grows with the
number  of objects.   So  far the  only recognition  approaches  that have  been
demonstrated to work on large scale  problems are based on Nearest Neighbor~(NN)
classification~\cite{Nister06,Jegou11,Dean13},   because   extremely   efficient
methods    for   NN    search   exist    with   an    average   complexity    of
$O(1)$~\cite{Norouzi14,Muja14}.  Moreover,  Nearest Neighbor~(NN) classification
also offers  the possibility to trivially  add new objects, or  remove old ones,
which is not  directly possible with neural networks, for  example.  However, to
the best  of our knowledge, such an approach has not  been applied to the  3D pose
estimation problem,  while it can potentially  scale to many objects  seen under
large  ranges of  poses.   For  example, \cite{Dean13}  only  focuses on  object
recognition without considering the 3D pose estimation problem.

For  NN approaches  to perform  well, a  compact and  discriminative description
vector is required.  Such representations that  can capture the appearance of an
object      under      a      certain      pose      have      already      been
proposed~\cite{Dalal05,Hinterstoisser12b}, however they  have been handcrafted.  Our
approach is motivated by the success  of recent work on feature point descriptor
learning~\cite{Brown10,Trzcinski13a,Masci14}, which shows that it is possible to
learn compact descriptors that significantly outperform handcrafted methods such
as SIFT or SURF.

\begin{figure*}[tbp]
\center
\includegraphics[width=0.99\textwidth]{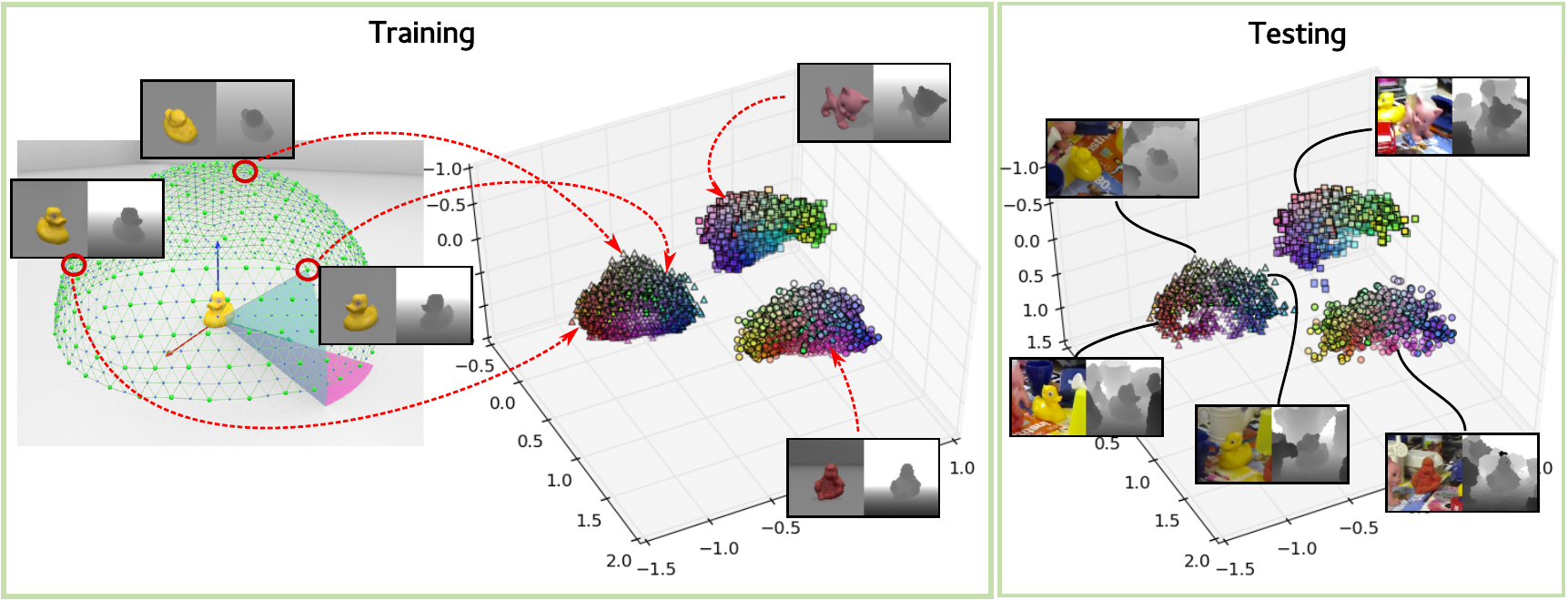} \\ \vspace{1mm}
\includegraphics[width=0.99\textwidth]{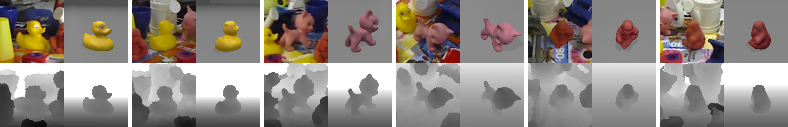}
\caption{Three-dimensional descriptors for several objects under many different
  views computed by our method on RGB-D data.  {\bf Top-left:} The training
  views of different  objects are mapped to well-separated  descriptors, and the
  views of the  same object are mapped to descriptors  that capture the geometry
  of  the  corresponding  poses,  even  in this  low  dimensional  space.   {\bf
    Top-right:} New images  are mapped to locations corresponding  to the object
  and 3D poses, even in the presence of clutter.  {\bf Bottom:} Test RGB-D views and
  the RGB-D data corresponding to the closest template descriptor.}
\label{fig:one}
\end{figure*}

However,  the problem  we  tackle  here is  more  complex:  While feature  point
descriptors are used only  to find the points' identities, we here want to find
both  the  object's  identity and  its  pose.   We  therefore  seek to  learn  a
descriptor with the two following  properties: a) The Euclidean distance between
descriptors  from  two different  objects  should  be  large; b)  The  Euclidean
distance between  descriptors from the  same object should be  representative of
the similarity between their  poses.  This way, given a new  object view, we can
recognize the object and get an estimate  of its pose by matching its descriptor
against a database of registered descriptors.  New objects can also be added and
existing ones removed easily.   To the best of our knowledge,  our method is the
first one that learns to compute descriptors for object views.

Our approach is related  to manifold learning, but  the key advantage of learning a direct mapping to descriptors is that  we can  use efficient  and scalable  Nearest Neighbor  search methods.   
This is  not  possible for  previous methods  relying on  geodesic
distances on manifolds.  Moreover,  while previous approaches already considered
similar  properties to  a) and  b), to  the best  of our  knowledge, they  never
considered  both  simultaneously,  while  it  is  critical  for  efficiency.
Combining these two constraints in a principled  way is far from trivial, but we
show it  can be  done by  training a Convolutional  Neural Network~\cite{LeCun98}  using simple
constraints   to   compute    the    descriptors.     As   shown    in
Fig.~\ref{fig:one}, this results in a method  that nicely untangles the views of
different objects into descriptors that capture  the identities and poses of the
objects.

We evaluate our  approach on instance recognition and pose  estimation data with
accurate  ground truth  and  show significantly  improved  results over  related
methods.   Additionally we  perform  experiments assessing  the  ability of  the
method to generalize to unseen objects showing promising results.

\section{Related Work}

Our work is related to several aspects of Computer Vision, and we focus here on the
most relevant and representative work.
Our   approach   is   clearly   in   the   framework   of   2D   view   specific
templates~\cite{Hoiem11},   which   is   conceptually   simple,   supported   by
psychophysical  experiments~\cite{Tarr89},  and   was  successfully  applied  to
various       problems       and        datasets       over       the       last
years~\cite{Nayar96b,Malisiewicz11,Hinterstoisser12,Gu10,Dean13,RiosCabrera14}.

However,  most  of  these  works  rely on  handcrafted  representations  of  the
templates,          for          example          HOG~\cite{Dalal05}          or
LineMOD~\cite{Hinterstoisser12b}. In particular,  LineMOD was designed explicitly
in  the  context  of  object  detection  and  pose  estimation.   However  these
handcrafted representations  are  suboptimal  compared to  statistically  learned  features.
\cite{Malisiewicz11,Gu10,RiosCabrera14}   show  how  to  build
discriminative  models based  on  these representations  using  SVM or  boosting
applied  to  training  data. \cite{Malisiewicz11,RiosCabrera14} do  not consider  the  pose
estimation  problem, while  \cite{Gu10} focuses  on  this problem  only, with  a
discriminatively trained mixture of HOG  templates. Exemplars were recently used
for 3D object detection and pose estimation in~\cite{Aubry14}, but still rely on
a handcrafted representation.

As mentioned in  the introduction, our work is influenced  by work developed for
keypoint descriptor  learning.  Some  of these methods  are applied  to existing
descriptors     to     make     them    more     discriminative,     such     as
in~\cite{Gong12,Strecha12},  but  others are  trained  directly  on image  data.
\cite{Brown10}  introduces  datasets  made  of  ``positive  pairs''  of  patches
corresponding  to the  same physical  points and  ``negative pairs''  of patches
corresponding to different points. It is used for example in~\cite{Trzcinski13a}
to  learn a  binary descriptor  with  boosting.  \cite{Masci14}  uses a  ``siamese''
architecture~\cite{Chopra05} to train a neural network to compute discriminative
descriptors. Our approach is  related to this last work, but  the notion of pose
is absent in their  case. We show how to introduce this  notion by using triplets
of training examples in addition to only pairs.

Instead  of  relying  on rigid  templates  as  we  do,  many works  on  category
recognition and  pose estimation  rely on part-based  models.  \cite{Savarese07}
pioneered this  approach, and learned canonical  parts connected by a  graph for
object recognition  and pose estimation.  \cite{Pepik12}  extends the Deformable
Part  Model to  3D object  detection and  pose estimation.   \cite{Payet11} uses
contours as parts. One major drawback of such approaches is that the complexity is
typically linear with the number of objects.  It is also not clear how important
the ``deformable'' property really is for the  recognition, and rigid templates seem to
be sufficient~\cite{Divvala12}.

Our   approach   is   also    related   to   manifold   learning~\cite{Pless09}.
For example, \cite{Salakhutdinov07}  learns an  embedding that  separates extremely  well the
classes  from the  MNIST dataset  of digit  images, but  the notion  of pose  is
absent.  \cite{Hadsell06} learns either for different classes, also on the MNIST
dataset, or for  varying pose and illumination, but not  the two simultaneously.
More recently,  \cite{Bakry14} proposes a  method that separates  manifolds from
different categories while being able to predict the object poses, and also does
not require  solving an  inference problem, which  is important  for efficiency.
However, it relies on a discretisation of the pose space in a few classes, which
limits the  possible accuracy.  It  also relies on  HOG for the  image features,
while we learn the relevant image features.

Finally, many works focus as we  do on instance recognition and pose estimation,
as    it    has    important    applications   in    robotics    for    example.
\cite{Hinterstoisser12b} introduced LineMOD, a fast but handcrafted presentation
of  template  for  dealing  with  poorly  textured  objects.   The  very  recent
\cite{Brachmann14,Tejani14}  do not  use templates  but rely  on recognition  of
local patches instead. However they were demonstrated on RGB-D images, and local
recognition is likely to be much more challenging  on poorly textured objects when a
depth information is not available.  \cite{Lai11} also expects RGB-D images, and
uses a tree for object recognition,  which however still scales linearly in the numbers of objects, categories, and poses.


\section{Method}

Given a  new input  image $x$  of an object,  we want  to correctly  predict the
object's class  and 3D pose.  Because  of the benefits discussed  above, such as
scalability  and ability  to easily  add and  remove objects,  we formulate  the
problem as a k-nearest neighbor search in a descriptor space: For each object in
the database, descriptors are calculated for  a set of template views and stored
along with the  object's identity and 3D pose  of the view.  In order  to get an
estimate for the class  and pose of the object depicted in  the new input image,
we  can  compute a  descriptor  for $x$  and  search for  the most  similar
descriptors in  the database. The output  is then the object  and pose associated
with them.

We therefore introduce a  method to efficiently map an input  image to a compact
and discriminative  descriptor that can be  used in the nearest  neighbor search
according to  the Euclidean distance.  For the  mapping, we use  a Convolutional
Neural Network (CNN) that is applied to the  raw image patch as  input and   delivers the
descriptor as activations of the last layer in one forward pass.

We show below  how to train such  a CNN to enforce the  two important properties
already  discussed  in  the  introduction: a)  The  Euclidean  distance  between
descriptors  from  two different  objects  should  be  large; b)  The  Euclidean
distance between  descriptors from the  same object should be  representative of
the similarity between their poses.

\newcommand{\Strain}{\mathcal{S}_\text{train}}
\newcommand{\Sdb}{\mathcal{S}_\text{db}}
\newcommand{\Pairs}{\mathcal{P}}
\newcommand{\Triplets}{\mathcal{T}}

\subsection{Training the CNN}

In order to train the network we need a set $\Strain$ of training samples, where
each sample $s = (x, c, p)$ is made of an image $x$ of an object, which can be a
color or grayscale image  or a depth  map, or a  combination of the  two; the
identity $c$ of  the object; and the 3D  pose $p$ of the object  relative to the
camera.

Additionally, we define a  set $\Sdb$ of templates where each element  is defined in the
same way as  a training sample.  Descriptors for these  templates are calculated
and stored with the classifier for  k-nearest neighbor search.  The template set
can be a subset  of the training set, the whole training set  or a separate set.
Details  for  the creation  of  training  and template  data  are  given in  the
implementation section.

\subsection{Defining the Cost Function}

We argue  that a good mapping  from the images  to the descriptors should  be so
that  the Euclidean  distance between  two descriptors  of the  same object  and
similar poses  are small and  in every other  case (either different  objects or
different poses) the  distance should be large.  In  particular, each descriptor
of a training sample should have a small distance to the one template descriptor
from the  same class with  the most  similar pose and  a larger distance  to all
descriptors of templates from other classes,  or the same class but less similar
pose.

\newcommand{\Fcnn}{f_{\!_w\!}}
\newcommand{\Lpairs}{\mathcal{L}_\text{pairs}}
\newcommand{\Ltriplets}{\mathcal{L}_\text{triplets}}

We enforce  these requirements  by minimizing  the following  objective function
over the parameters $w$ of the CNN:
\begin{equation}
\mathcal{L} = \Ltriplets + \Lpairs + \lambda {||w'||}_2^2  \;\; .
\end{equation}
The  last term  is a  regularization term  over the  parameters of  the network:
$w'$ denotes  the vector  made of  all the  weights of  the convolutional
filters and all nodes of the fully connect layers, except the bias terms. We
describe the first two terms $\Ltriplets$ and $\Lpairs$ below.

\subsubsection{Triplet-wise terms}
\label{sec:triplet_cost}
We first  define a  set $\Triplets$  of triplets $(s_i,  s_j, s_k)$  of training
samples.   Each triplet  in $\Triplets$  is  selected such that  one of  the two  following conditions is fulfilled:
\begin{itemize}
\item either $s_i$ and $s_j$ are from the same object and $s_k$ from another object, or
\item the three  samples $s_i$, $s_j$, and  $s_k$ are from the  same object, but
  the poses $p_i$  and $p_j$ are more similar than the  poses $p_i$ and
  $p_k$.
\end{itemize}
These triplets can therefore be seen as made of a pair of similar samples ($s_i$
and $s_j$) and a pair of dissimilar  ones ($s_i$ and $s_k$). We introduce a cost
function for such a triplet:
\begin{align}
c(s_i, s_j, s_k) = \max \left( 0, 1 - \frac{{||\Fcnn(x_i) - \Fcnn(x_k)||}_2}{{||\Fcnn(x_i) - \Fcnn(x_j)||}_2 + m} \right) \;\; ,
\label{eq:triplet_cost}
\end{align}
where $\Fcnn(x)$ is  the output of the CNN  for an input image $x$  and thus our
descriptor for $x$, and $m$ is a margin. We can now define the
term $\Ltriplets$  as the  sum of this  cost function over  all the  triplets in
$\Triplets$:
\begin{equation}
\Ltriplets = \sum_{(s_i,  s_j, s_k) \in \Triplets} c(s_i, s_j, s_k) \;\; .
\end{equation}
It  is easy  to  check that  minimizing $\Ltriplets$  enforces  our two  desired
properties in one common framework.

The  margin $m$  serves two  purposes. First,  it introduces  a margin  for the
classification. It also  defines a minimum ratio for the  Euclidean distances of
the  dissimilar  pair  of samples  and  the  similar  one.   This
counterbalances the weight  regularization term,  which naturally  contracts the
output of the network and thus the descriptor space. We set $m$ to $0.01$ in all
our experiments.

The concept of forming triplets from similar and dissimilar pairs is adopted from the field of metric learning, in particular, the method of~\cite{Weinberger2008}, where it is used to learn a Mahalanobis distance metric.
Note also that  our definition of the cost is slightly different from the one in~\cite{Wang14}, which uses $c(s_i, s_j, s_k)  = \max
\left(   0,  m   +   {||\Fcnn(x_i)  -   \Fcnn(x_j)||}_2^2   -  {||\Fcnn(x_i)   -
  \Fcnn(x_k)||}_2^2 \right)$, where $m$ is set to 1.  
Our formulation does not suffer from a vanishing gradient when the distance of the dissimilar pair is very small (see suppl.\ material). Also the increase of the cost with the distance of the similar pair is bounded, thus putting more focus on local interactions.
In practice, however, with proper initialization and selection of $m$ both formulations deliver similar results.

\subsubsection{Pair-wise terms}

In addition to the triplet-wise terms, we also use pair-wise terms. These terms
make the descriptor robust to noise and other distracting artifacts such as changing illumination. We consider
the set $\Pairs$ of pairs $(s_i, s_j)$ of samples from the same object under
very similar poses, ideally the same, and we define the $\Lpairs$ term as the sum of the squared
Euclidean distances between the descriptors for these samples:
\begin{equation}
\Lpairs = \sum_{(s_i,  s_j) \in \Pairs} {||\Fcnn(x_i) - \Fcnn(x_j)||}_2^2 \;\; .
\end{equation}

This term therefore enforces the fact that for two images of the same object and
same pose, we want to obtain two descriptors which are as close as possible to
each other, even if they are from different imaging conditions: Ideally we want
the same descriptors even if the two images have different backgrounds or
different illuminations, for example. As  will be discussed in more detail in
Section~\ref{sec:dataset}, this also allows us to use a mixture of real and synthetic  images for training.

Note  that  we  do  not  consider  dissimilar  pairs  unlike  work  in  keypoint
descriptors learning for example.  With  dissimilar pairs the problem arises how
strong  to penalize  a certain  distance between  the two  samples, given  their
individual labels.  Using triplets instead gives the  possibility to
only consider relative dissimilarity.

\subsection{Implementation Aspects}
\label{sec:impl_aspects}

\begin{figure}[tbp]
\center
\includegraphics[width=0.98\columnwidth]{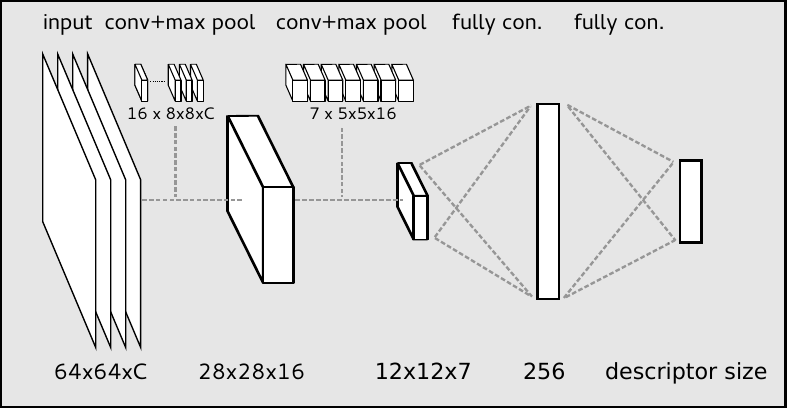}%
\caption{Network structure:  We use a CNN made  of two convolutional  layers with
  subsequent  $2\times 2$  max  pooling layers,  and two  fully  connected layers.  The
  activations of the last layer form the descriptor for the input image.}
\label{fig:network_structure}%
\end{figure}

The exact structure of the network we  train to compute the descriptors is shown
in Figure~\ref{fig:network_structure}.   It consists of two  layers that perform
convolution of  the input with  a set  of filters, max-pooling  and sub-sampling
over  a $2\times  2$ area  and a  rectified linear  (ReLU) activation  function,
followed by  two fully connected layers.   The first fully connected  layer also
employs a  ReLU activation, the  last layer has  linear output and  delivers the
final descriptor.

We optimize the parameters $w$ of the CNN by Stochastic Gradient Descent on
mini-batches     with    Nesterov     momentum~\cite{Sutskever2013ICML}.     Our
implementation is based on Theano~\cite{bergstra2010scipy}.

The implementation  of the optimization  needs some  special care: Since  we are
working with mini-batches, the data corresponding to each pair or triplet has to
be organized such as to reside  within one mini-batch.  
The most straightforward implementation would be to place the data for each pair and triplet after each other, calculate the resulting gradient wrt.\ the network's parameters individually and sum them up over the mini-batch.
However, this  would  be  inefficient since  descriptors for templates would be  calculated multiple times if they appear in more than one pair or triplet. 

To assemble a mini-batch we start by randomly  taking one training sample from each  object. Additionally, for each of them we add its template with the most similar pose, unless it was already included in this mini-batch.  
This is iterated until the mini-batch is full. 
However, this procedure can lead to very unequal numbers of templates per object if, for instance, all of the selected training samples have the same most similar template. 
We make  sure that for each object at least two templates are included by adding a random one if necessary.
Pairs are then formed by associating each training sample  with its  closest template.   Additionally, for each   training sample in the mini-batch
we initially create three triplets. In each of  them the similar template is  set to be the one with the closest pose and the dissimilar sample is either another, less similar template from the same object or any template of a different object.

During  the optimization,  after the  first set of epochs,  we perform  boot-strapping of
triplets within  each mini-batch  to focus  on the  difficult samples:  For each
training sample  we add two additional  triplets.  The similar template is again the closest one.  The  dissimilar ones are those templates  that currently have
the closest  descriptors, one from  the same object  but different pose  and one
from all the other objects.

Another aspect to  take  care of  is the fact  that the  objective function  must be
differentiable with respect  to the parameters of the CNN,  while the derivative
of  the square  root---used  in the  triplet-wise cost---is  not  defined for  a
distance of 0. Our solution is to  add a small constant $\epsilon$ before taking
the square root.  Another possible  approach~\cite{Wang14} is to take the square
of the norm. However, this induces the problem (mentioned in Section~\ref{sec:triplet_cost}) that for  very small  distances of  the dissimilar  pair, the  gradient becomes very small and vanishes for zero distance.


\section{Evaluation}

We   compare    our   approach   to    LineMOD   and   HOG   on    the   LineMOD
dataset~\cite{Hinterstoisser12}.  This  dataset contains training and  test data
for object recognition  and pose estimation of 15 objects,  with accurate ground
truth.  It comes with  a 3D  mesh for each  of the objects.   Additionally, it also
provides  sequences of  RGB  images and  depth maps  recorded  with a  \Kinect{}
sensor.

\subsection{Dataset Compilation}
\label{sec:dataset}

We train a CNN  using our method on a mixture of synthetic  and real world data.
As in~\cite{Hinterstoisser12b},  we create synthetic training  data by rendering
the mesh available  for each of the  objects in the dataset from  positions on a
half-dome over  the object,  as shown  in Fig.~\ref{fig:one}  on the  left.  The
viewpoints are  defined by starting  with a regular icosahedron  and recursively
subdividing each triangle into 4  sub-triangles.  For the template positions the
subdivison is applied two times.  After removing the lower half-sphere we end up
with 301  evenly distributed  template positions.   Additional training  data is
created by subdividing one more time, resulting in 1241 positions.

From each pose we render the object standing on a plane over an empty background
using Blender\footnote{http://www.blender.org}.  We parameterize the object pose
with the azimuth  and elevation of the  camera relative to the  object. We store
the RGB image as well as the depth  map.

For  the real  world data  we  split the  provided sequences  captured with  the
\Kinect{}  randomly  into  a  training  and  a test  set.   We  ensure  an  even
distribution of the samples over the viewing hemisphere by taking two real world
images close  to each template,  which results roughly in  a 50/50 split  of the
data into training  and test.  Preliminary experiments showed very  little to no
variance over the  different train/test splits and, thus,  all results presented
here report runs on one random split, fixed for each experiment.

The whole  training data set is  augmented by making multiple  copies with added
noise. On both  RGB and depth channel  we add a small amount  of Gaussian noise.
Additionally, for the synthetic images, we  add larger fractal noise
on the background,  to simulate diverse backgrounds.

Note that  the template views, which  are ultimately used in  the classification
are  purely  synthetic and  noise-free  renderings  on clean  backgrounds.   The
algorithm, thus, has to learn to map the  noisy and real world input data to the
same location in descriptor space as the clean templates.

As pointed out in~\cite{Hinterstoisser12b} some  of the objects are rotationally
invariant, to different degrees.  Thus, the  measure of similarity of poses used
for the  evaluation and, in  our case to define  pairs and triplets,  should not
consider  the azimuth  of the  viewing  angle for  those objects.  We treat  the
\emph{bowl} object  as fully  rotationally invariant. The  classes \emph{eggbox},
\emph{glue} are treated  as symmetric, meaning a rotation  by $180^\circ$ around
the z-axis shows the  same pose again. The \emph{cup} is  a special case because
it looks  the same from  a small range of  poses, but from  sufficient elevation
such that  the handle is  visible, the exact pose  could be estimated.   We also
treat it  as rotationally invariant,  mainly to  keep the comparison  to LineMOD
fair.

We extract a patch  centered at the object and capturing a  fixed size window in
3D  at the  distance of  the  object's center.   In  order to  also address  the
detection part in a sliding window manner,  it would be necessary to extract and
test  several  scales.  However, only  a  small  range  of  scales needs  to  be
considered, starting  with a  maximal one,  defined by the  depth at  the center
point, and  going down until  the center of the  object is reached.

Before  applying  the  CNN  we  normalize the  input  images.   RGB  images  are
normalized to the usual zero mean, unit variance. For depth maps we subtract the
depth at  the center of  the object,  scale down such  that $20$cm in  front and
behind  the  object's center  are  mapped  to the  range  of  $[-1,1]$ and  clip
everything beyond that range.

The test sequences  captured with the \Kinect{} are very  noisy.  In particular,
there are  many regions with undefined  depth, introducing very large  jumps for
which  the convolutional  filters with  ReLU activation  functions might  output
overly  strong  output values.   Therefore,  we  pre-process  the test  data  by
iteratively applying median  filters in a $3\times 3$ neighborhood,  but only on
the pixels for which the depth is available, until all gaps are closed.

\subsection{Network Optimization}

For  the optimization  we use  the following  protocol: We  initially train  the
network on  the initial dataset for  400 epochs, an initial learning rate of $0.01$ and a momentum of 0.9.  Every 100  epochs the learning rate  is multiplied  by  $0.9$.  Then  we perform  two  rounds of  bootstrapping triplet  indices as  explained in  Section~\ref{sec:impl_aspects}, and  for each round we train the CNN for another 200 epochs on the augmented training set.  In the end we train  another 300 epochs with the learning rate  divided by $10$ for final fine-tuning. The  regularization weight $\lambda$ is set to $10^{-6}$ in all our experiments.

\subsection{LineMOD and HOG}

We compare  our learned  descriptors to  the LineMOD  descriptor and  HOG as a baseline, as it is widely used as representation in the related work.  For
LineMOD we use the  publicly available source code in OpenCV.  We  run it on the
same  data as  our method,  except for  the median  filter depth  inpainting and
normalization:  LineMOD  handles the  missing  values  internally and  performed
better without these pre-processing operations.

For HOG we also use the  publicly available implementation in OpenCV. We extract
the HOG descriptors  from the same data we  use with our CNN. We  use a standard
setup of a $64 \times 64$ window size,  $8 \times 8$ cells, $2\times 2$ cells in
a block  and a  block stride  of $8$, giving  a 1764-dimensional  descriptor per
channel. We compute  descriptors on each RGB and depth  channel individually and
stack them.  For evaluation we normalize  all descriptors to length $1$ and take
the dot product between test and template descriptors as similarity measure.

\begin{figure*}[tbp]
\center
\subfloat[ours]{
\includegraphics[width=0.32\textwidth]{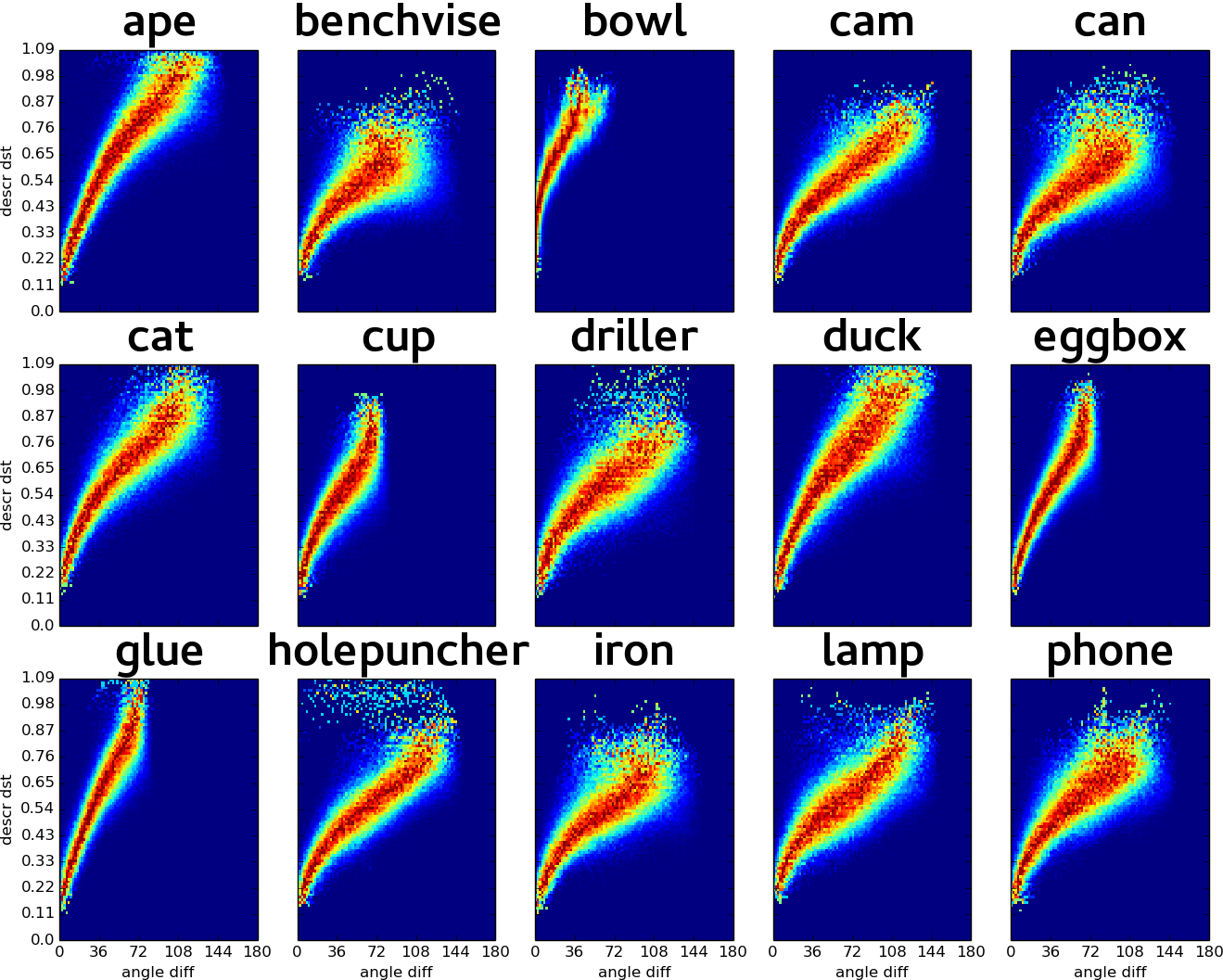} }%
\subfloat[LineMOD]{
\includegraphics[width=0.32\textwidth]{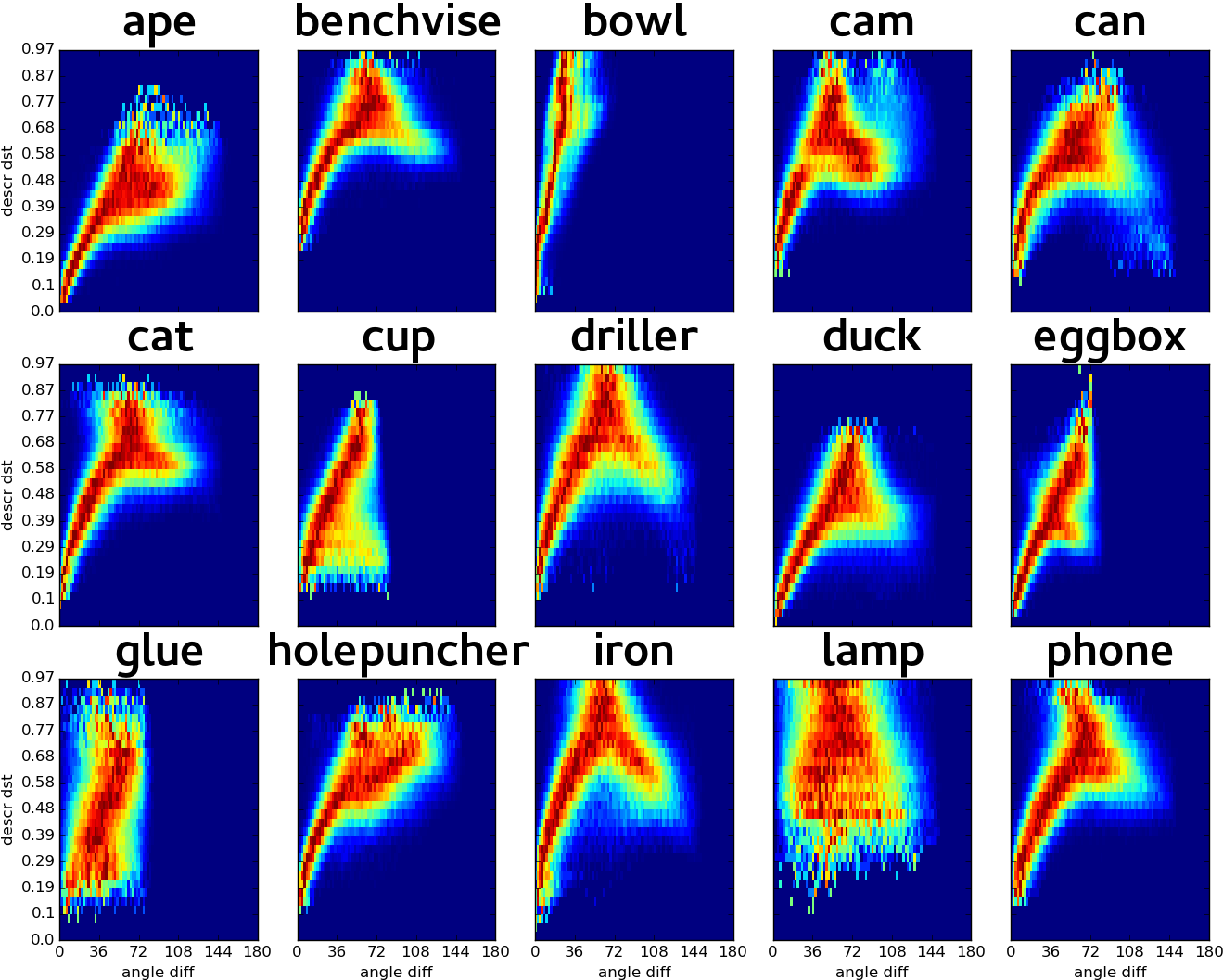} }%
\subfloat[HOG]{
\includegraphics[width=0.32\textwidth]{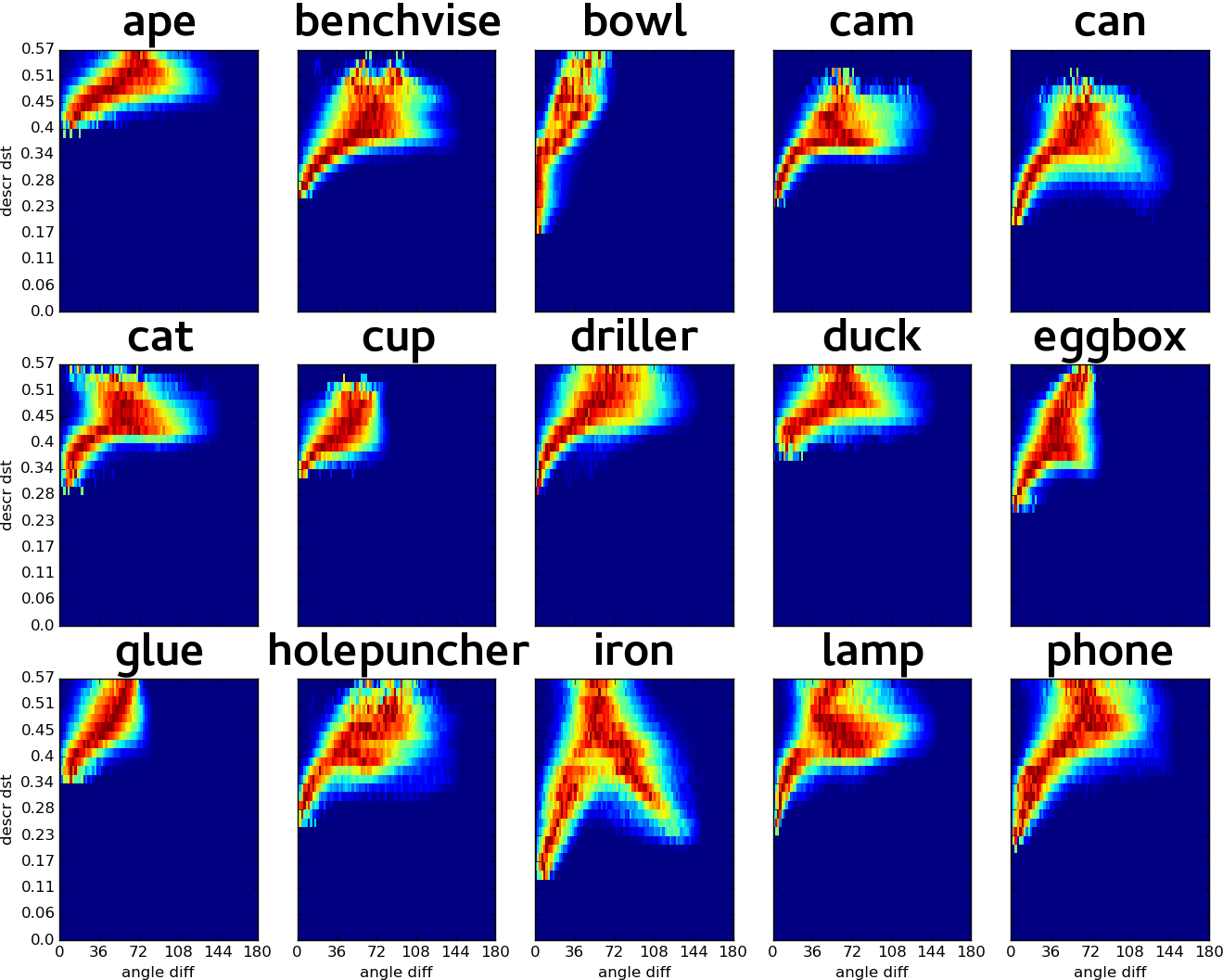} }%
\caption{Histograms of  the correlations between Pose  Similarity (x-axis) and Descriptor Distance (y-axis) for each of  the 15 objects of the LineMOD dataset  on RGB-D data, as
  described in  Section~\ref{sec:manifolds}.  Distances in the  descriptor space
  are much more  representative of the similarity between poses  with our method
  than with LineMOD or HOG. }
\label{fig:sim_vs_dst}
\end{figure*}

\begin{figure*}[tbp]
\center
\subfloat[depth]{
\includegraphics[width=0.32\textwidth,height=0.2\textwidth,trim=1cm 5mm 1cm 1cm]{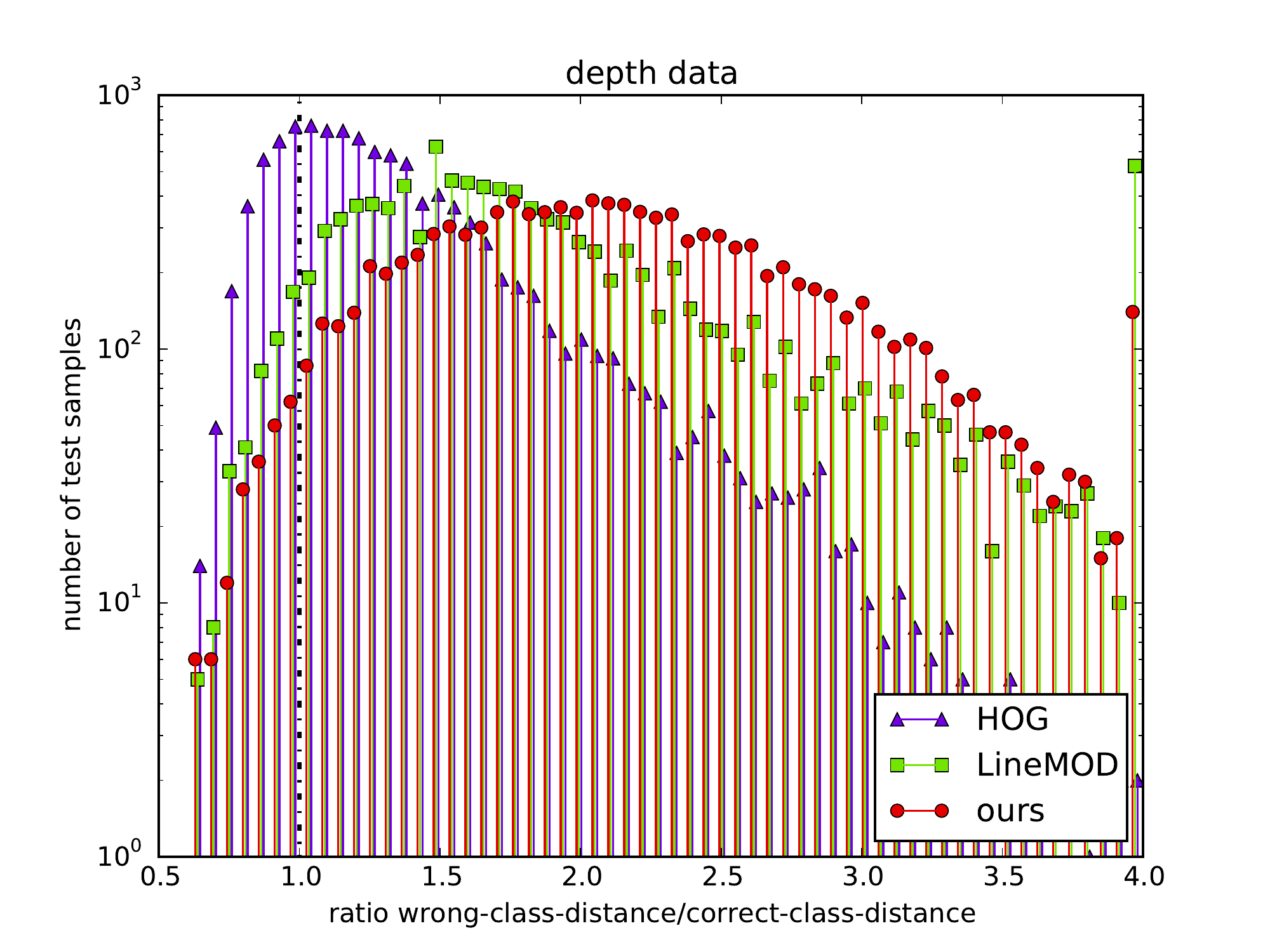} }%
\subfloat[RGB]{
\includegraphics[width=0.32\textwidth,height=0.2\textwidth,trim=1cm 5mm 1cm 1cm]{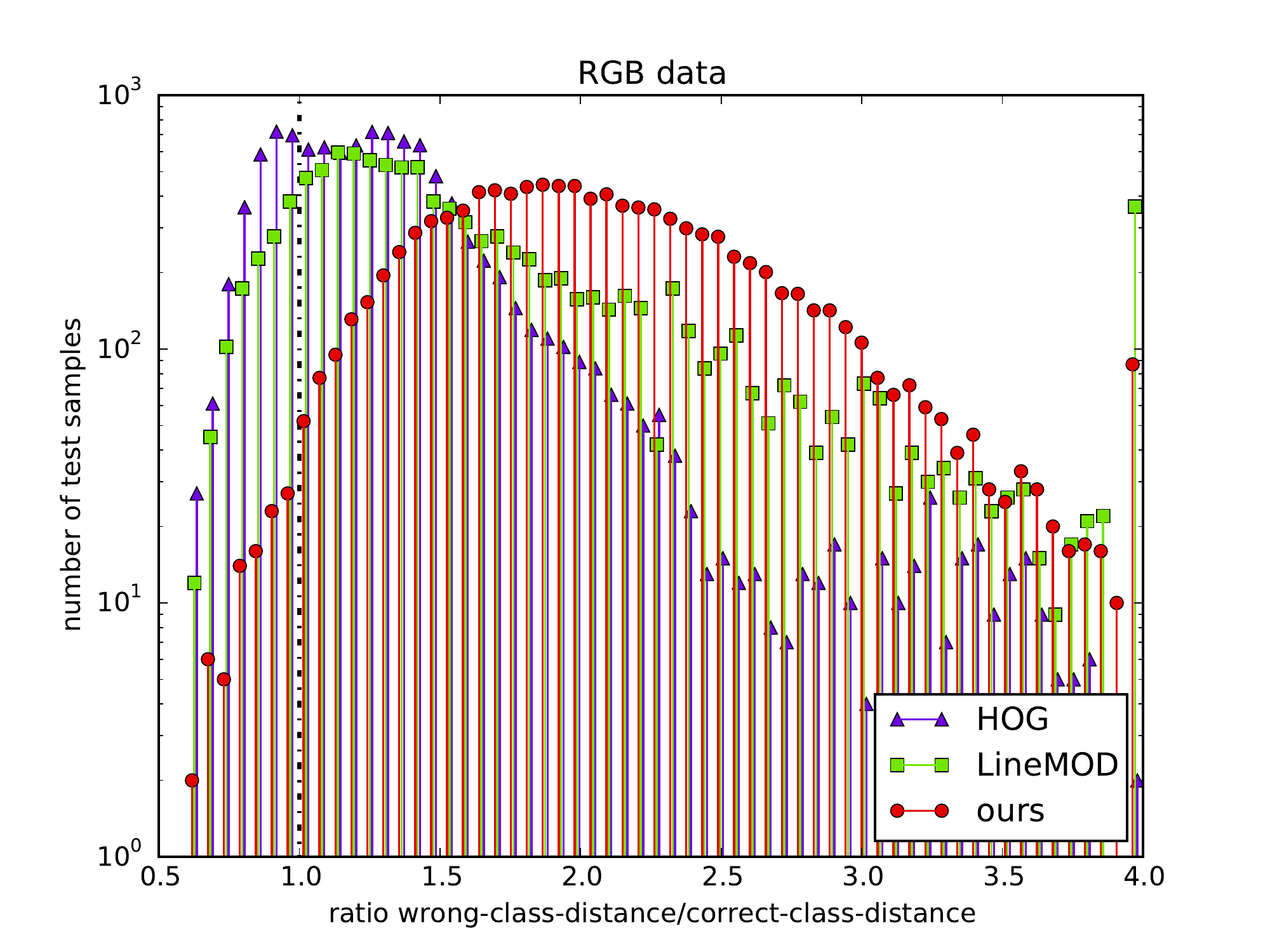} }%
\subfloat[RGB-D]{
\includegraphics[width=0.32\textwidth,height=0.2\textwidth,trim=1cm 5mm 1cm 1cm]{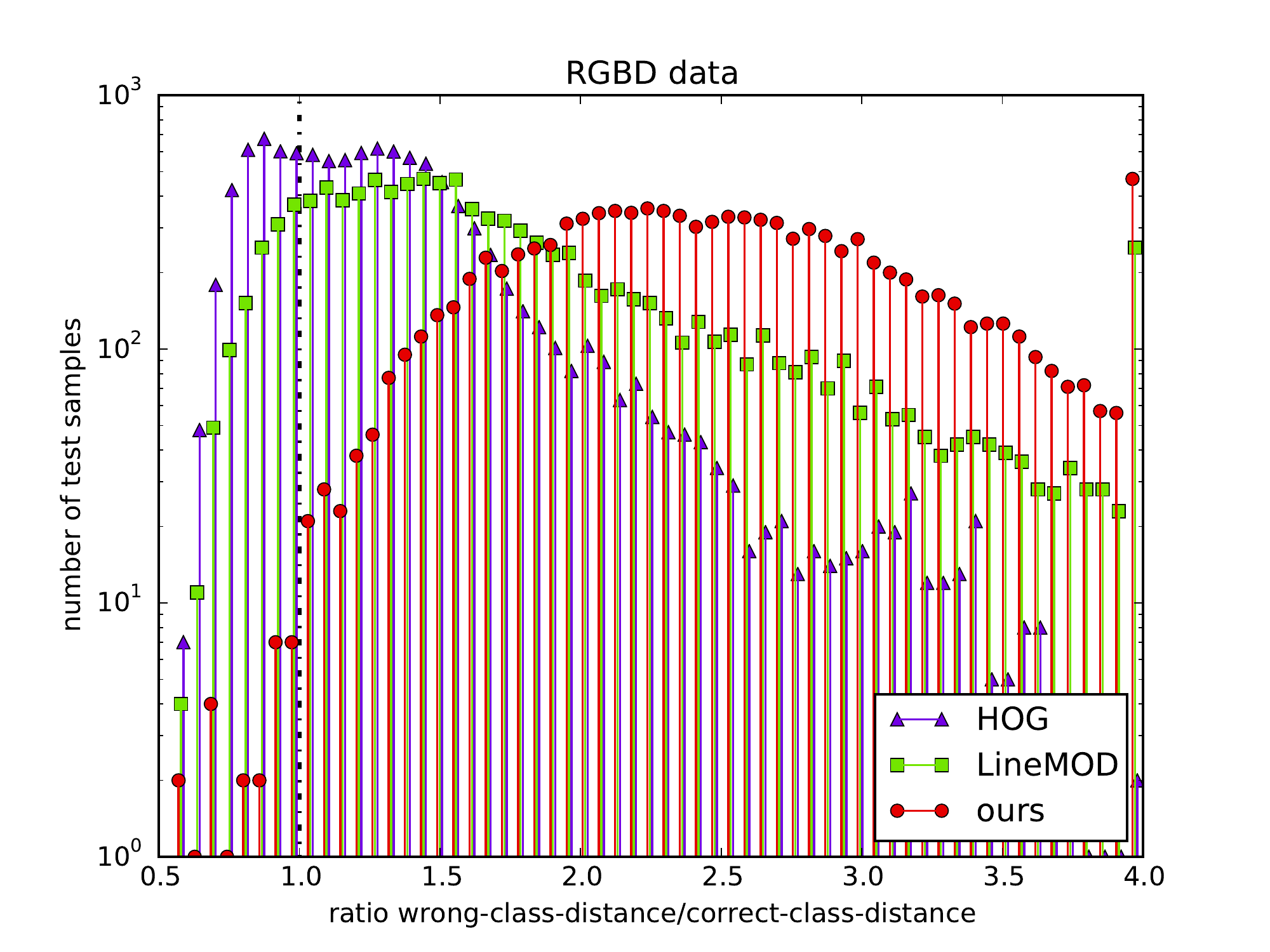} }%
\caption{Evaluation of the class separation, over  the 15 objects of the LineMOD
  dataset.  The histograms  plot the ratios between the distance  to the closest
  template from the correct object and the distance to the closest template from
  any other object, for depth, RGB, and  RGB-D data. Ratios above 4 are clipped;
  the y-axis is  scaled logarithmically. Our method has considerably fewer  samples for which
  the ratio is below one, which indicates less confusion between the objects.  }
\label{fig:class_sep}
\end{figure*}

\subsection{Manifolds}
\label{sec:manifolds}

Figure~\ref{fig:one} plots  the views of  three objects after being  mapped into
3-dimensional descriptors, for  visualization purposes.  As can  be seen, not
only the  descriptors from the  different objects  are very well  separated, but
they also capture the geometry of  the corresponding poses.  This means that the
distances between  descriptors is  representative of  the distances  between the
corresponding poses, as desired.

For longer descriptors we show an evaluation of the relation between distances of descriptors and similarity between poses in Figure~\ref{fig:sim_vs_dst}. For each object, we computed the distances between  every sample in the test set and every template for  the same object in  the training set, as well  as the angles between  their poses.   We  then  plot a  two-dimensional  histogram over  these angle/distance  pairs.  Correlation  between  small angles  and large  distances indicates the  risk of missed target templates,  and correlation between large  angles and small distances  the risk  of incorrect matches.  Ideally the  histograms should therefore have large values only on the diagonal. 

The histograms  for the descriptors computed  with our method clearly  show that
the distance between  descriptors increase with the angle between  the views, as
desired, while  the histograms for LineMOD  and HOG show that  these descriptors
are much more ambiguous.

Additionally, the ability  of the descriptors to separate  the different classes
is evaluated  in Figure~\ref{fig:class_sep}.  For  every test sample descriptor we compute the distance to the closest template descriptor of the same object and the closest from any
other  object and  plot  a  histogram over  those  ratios. Clearly,  descriptors
obtained  with our  method exhibit  a  larger ratio  for most  samples and  thus
separate the objects better.

\subsection{Retrieval Performance}

What we  ultimately want from the  descriptors is that   nearest neighbors are
from the same class and have similar  poses. In order to evaluate the performance
we thus perform the following comparisons.

The scores reported for LineMOD in \cite{Hinterstoisser12b} represent the accuracy of the output of the
whole processing pipeline, including the descriptor calculation, retrieval of similar templates, pruning the set with heuristics and refinement of the pose for a set of candidate matches by aligning a voxel model  of the object. 
The contribution of this work is to replace the descriptors for the retrieval of templates with similar pose. Thus, we evaluate and compare this step in separation of the rest of the pipeline.

\paragraph*{Evaluation Metric}

For each  test sample  we consider  the $k$-nearest  neighbors according  to the
descriptors and similarity metric of each  method, the Euclidean distance in our
case, the dot  product for HOG, and  the matching score of  LineMOD. Among those
$k$ nearest templates  we search for the  one with the best closest  pose to the
test sample's pose, assuming that this  one would perform best in the subsequent
refinement process and thus finally be  selected.  The pose error is measured by
the angle between  the two positions on the viewing  half-sphere.  We define the
accuracy as  the percentage  of test images  for which the  best angle  error is
below a certain  threshold.  The minimum angle error for  which perfect accuracy
can theoretically be reached is $5^\circ$,  because that is the maximal distance
of a test image to its closest template.

\paragraph*{Descriptor Length}

\begin{figure}[tbp]
\center
\includegraphics[width=0.9\columnwidth,height=0.63\columnwidth,trim=1cm 5mm 1cm 1cm]{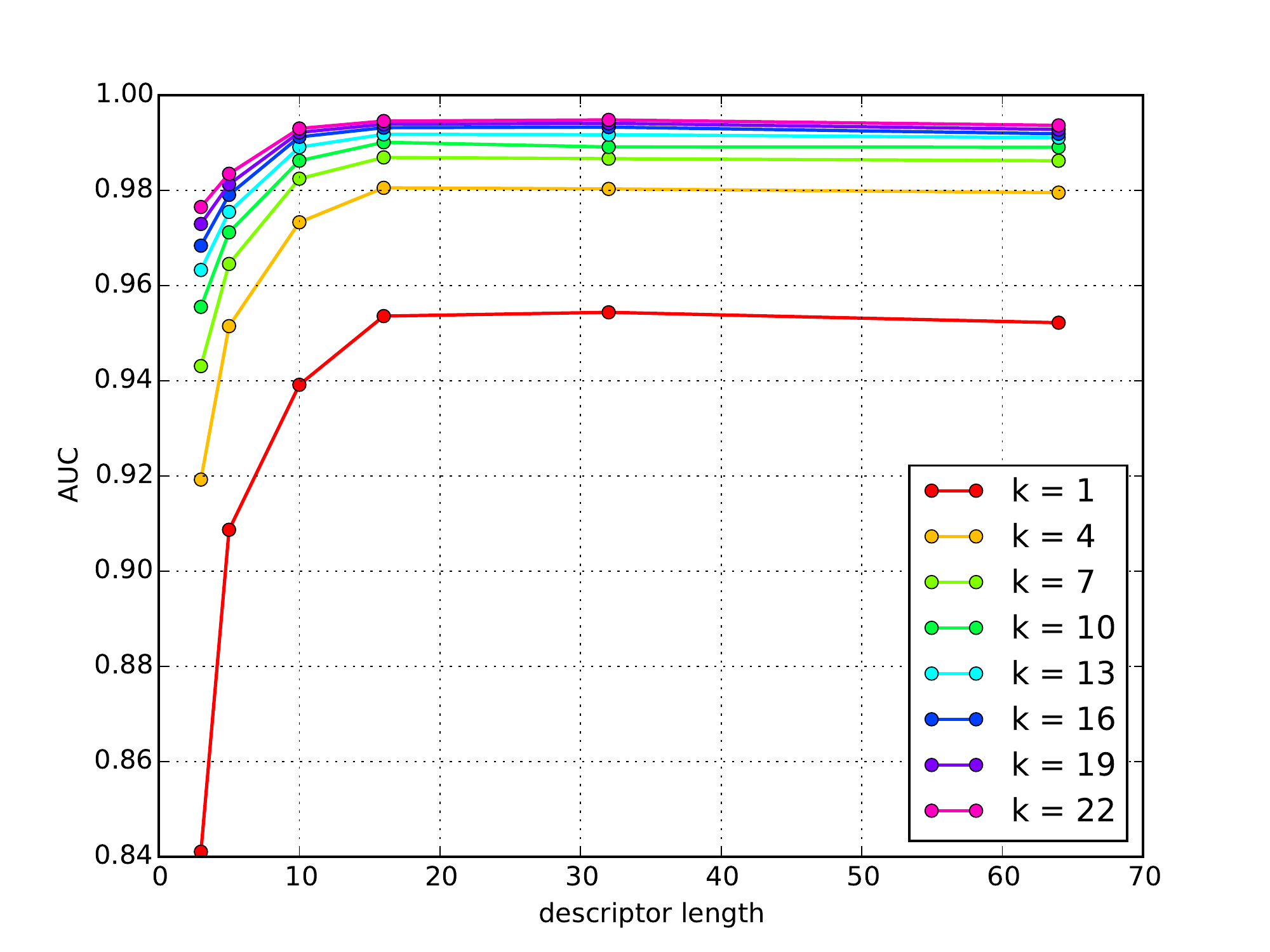}%
\caption{Evaluation of the  descriptor length on depth  data.  Only 16
  values are  sufficient to reliably  represent an object  view, after
  which the performance plateaus.  }
\label{fig:descr_len}
\end{figure}

In Figure~\ref{fig:descr_len}  we evaluate  the influence of  the length  of the
descriptors learned on depth data. As can be seen the maximal performance is
already reached  with a 16 dimensional  descriptor, while the length  of the HOG
descriptor  is 1764.   Thus,  we use a 16 dimensional  descriptor for  all  of the  following
experiments, including for the RGB and RGB-D data.

\paragraph*{Results}

\begin{figure*}%
\subfloat[depth]{ \label{fig:compare_o15_dpt}
\includegraphics[width=0.32\textwidth,trim=1cm 5mm 1cm 1cm]{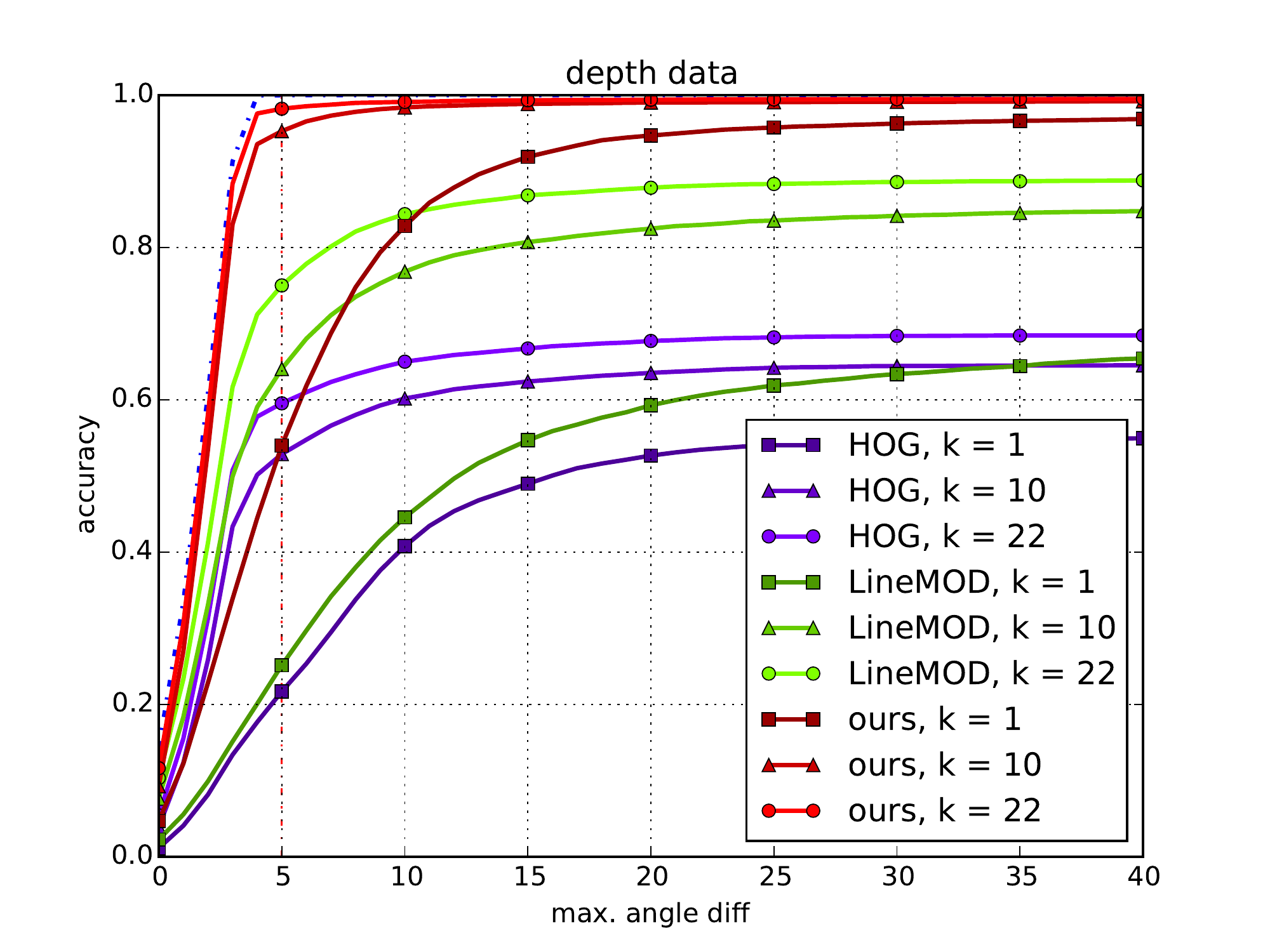} } %
\subfloat[RGB]{ \label{fig:compare_o15_rgb}
\includegraphics[width=0.32\textwidth,trim=1cm 5mm 1cm 1cm]{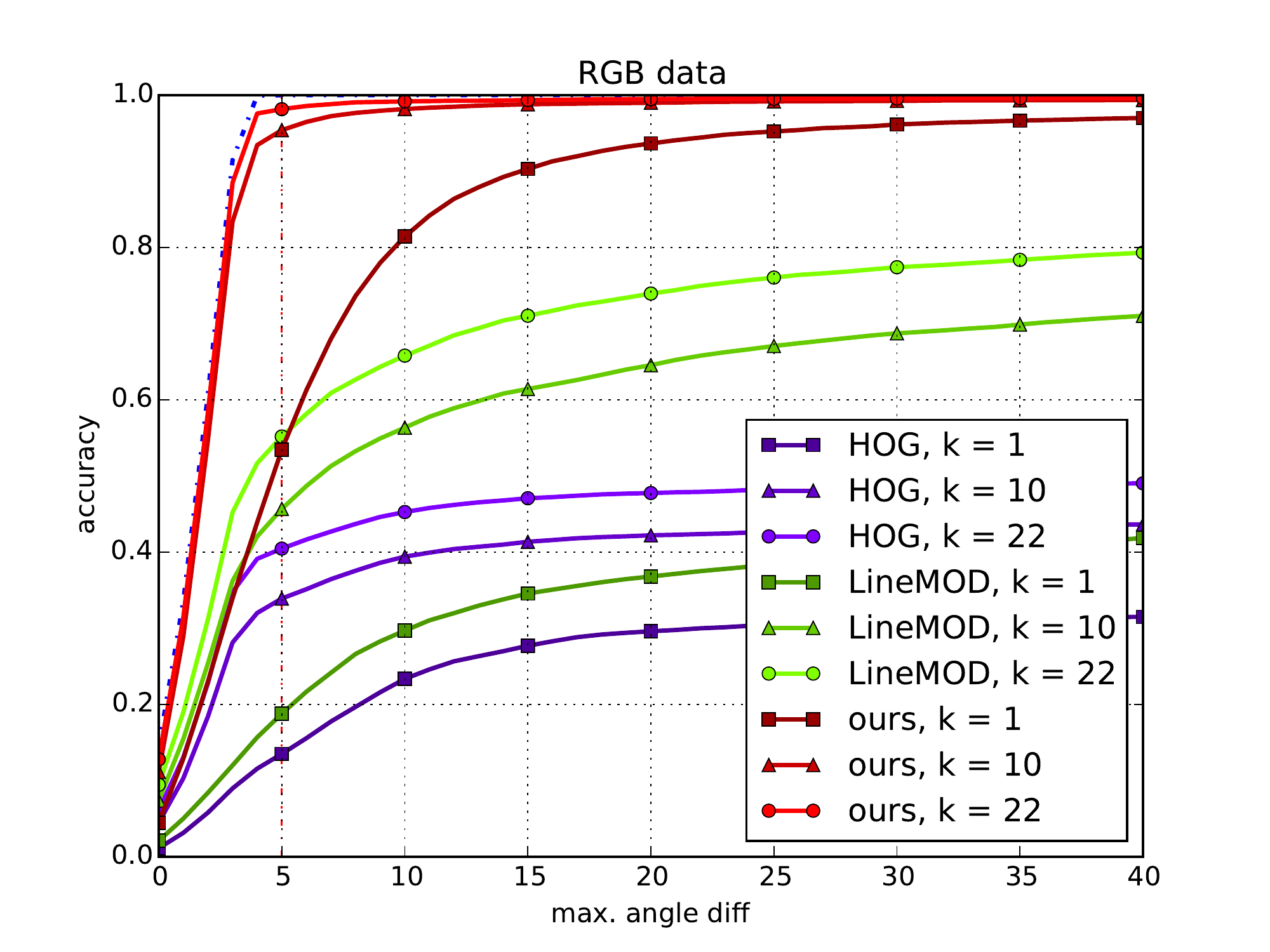} } %
\subfloat[RGB-D]{ \label{fig:compare_o15_rgbd}
\includegraphics[width=0.32\textwidth,trim=1cm 5mm 1cm 1cm]{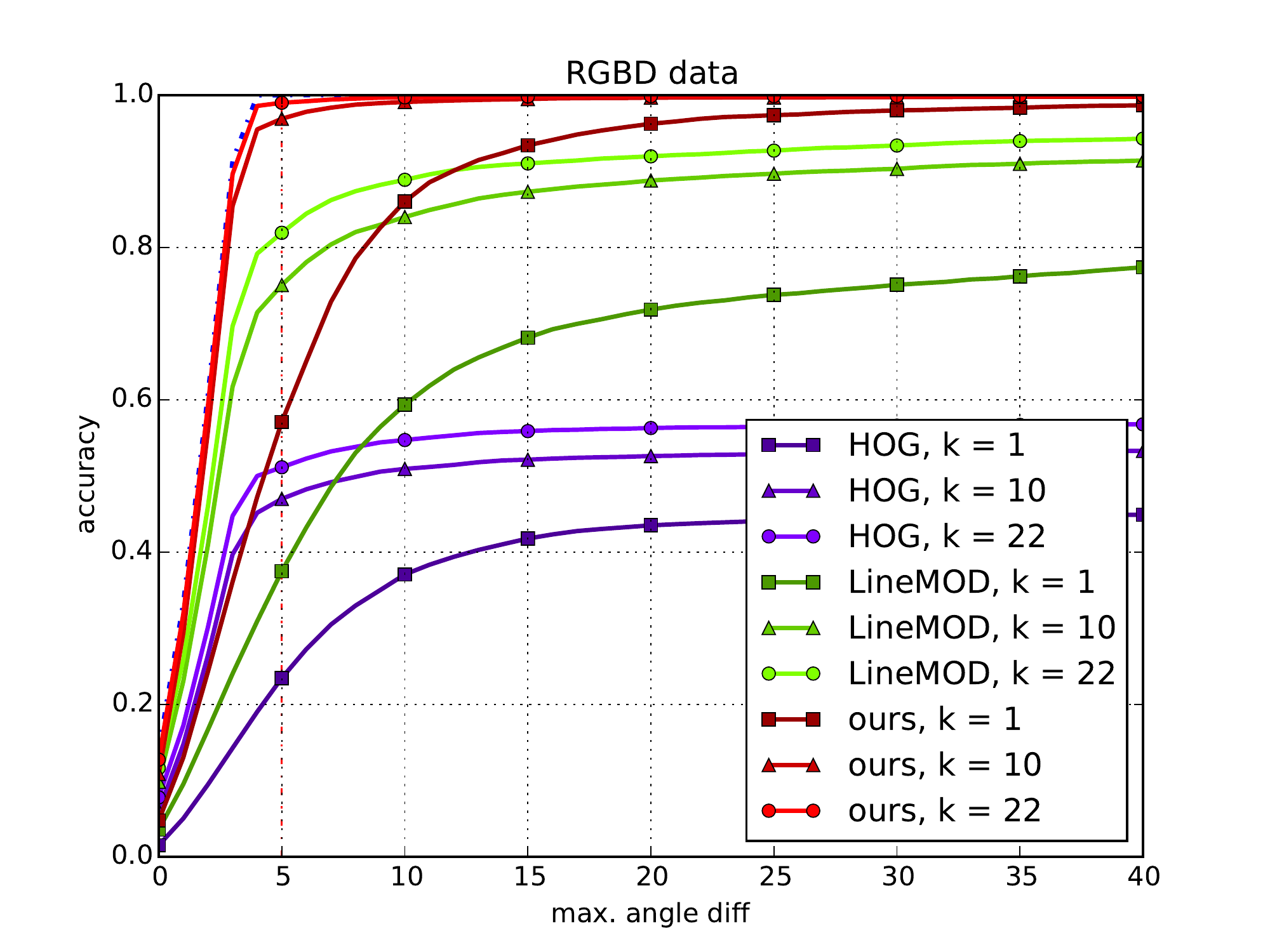} }%
\caption{Performance  evaluation and  comparison to  LineMOD and  HOG on  all 15
  objects of  the LineMOD dataset  and depth, RGB,  and RGB-D data.   Each graph
  plots the accuracy  over the y-axis for  given a maximal allowed  error of the
  resulting object's pose  on the x-axis. Curves for different  $k$ are computed
  by taking  k-nearest neighbors and  selecting the  one with the  best matching
  pose.  For our method,  the descriptor length was set to  $32$ for depth, RGB,
  and RGB-D data. HOG uses 1764 values per channel, and LineMOD uses a length of 63 for
  depth and RGB data, and of 126 for RGB-D data.}
\label{fig:compare_o15}
\end{figure*}

\begin{table*}[htb]
\small
\centering
\begin{tabular}{lcrrrrrrrrrrrr}
\toprule
 &  & \multicolumn{4}{c}{Depth} & \multicolumn{4}{c}{RGB} & \multicolumn{4}{c}{RGB-D} \\
 & k & $5^\circ$ & $20^\circ$ & $40^\circ$ & $180^\circ$ & $5^\circ$ & $20^\circ$ & $40^\circ$ & $180^\circ$ & $5^\circ$ & $20^\circ$ & $40^\circ$ & $180^\circ$ \\
\midrule
ours &     1 & {\bf 54.4} & {\bf 94.7} & {\bf 96.9} & {\bf 98.1}   & {\bf 53.4} & {\bf 93.7} & {\bf 97.0} & {\bf 99.1}  & {\bf 57.1} & {\bf 96.2} & {\bf 98.7} & {\bf 99.8} \\
LineMOD &  1 & 25.1 & 59.3 & 65.4 & 69.5   & 18.8 & 36.8 & 41.9 & 49.6  & 37.5 & 71.8 & 77.4 & 83.7 \\
HOG &      1 & 21.7 & 52.7 & 54.9 & 55.3   & 13.5 & 29.6 & 31.5 & 33.6  & 23.5 & 43.5 & 44.9 & 46.2 \\
\midrule
ours &    22 & {\bf 98.2} & {\bf 99.4} & {\bf 99.5} & {\bf 99.6}   & {\bf 98.2} & {\bf 99.5} & {\bf 99.6} & {\bf 99.7}  & {\bf 99.0} & {\bf 99.9} & {\bf 99.9} & {\bf 99.9} \\
LineMOD & 22 & 75.0 & 87.9 & 88.8 & 89.5   & 55.2 & 74.0 & 79.3 & 83.5  & 81.9 & 92.0 & 94.3 & 96.4 \\
HOG &     22 & 59.5 & 67.7 & 68.5 & 68.9   & 40.5 & 47.8 & 49.0 & 50.9  & 51.1 & 56.3 & 56.8 & 57.5 \\
\bottomrule
\end{tabular}
\vspace{1mm}
\caption{Performance comparison  to LineMOD  and HOG  on all  15 objects  of the
  LineMOD dataset  and depth, RGB, and  RGB-D data, for several  tolerated angle
  errors. Our  method systematically outperforms the  other representations. The
  value at  $180^\circ$ indicates the object  recognition rate when the  pose is
  ignored.}
\label{tab:angleerror_o15}
\end{table*}

We   evaluate  all   three   approaches   on  depth,   RGB,   and  RGB-D   data.
Figure~\ref{fig:compare_o15}  and  Table~\ref{tab:angleerror_o15} summarize  the
results. 
For \emph{depth maps}, results are shown in
Figure~\ref{fig:compare_o15}~\subref{fig:compare_o15_dpt}.        When      only
considering 1  nearest neighbor we  achieve a  recognition rate of  $98.1\%$, as
opposed to the $69.5\%$ achieved by the LineMOD descriptor and a pose error of less than $20^\circ$ for $94.7\%$ of the test samples ($59.3\%$ for LineMOD).
Figure~\ref{fig:compare_o15}~\subref{fig:compare_o15_rgb}   shows  results   for
training and testing on \emph{color images}.  While both LineMOD  and HOG cannot reach the  performance they  obtain on  the depth  data on  RGB alone,  our descriptor performs almost identically in this setup.
Finally, Figure~\ref{fig:compare_o15}~\subref{fig:compare_o15_rgbd} shows results
for training  and testing  on the  combination of color  images and  depth maps.
While LineMOD  takes advantage of the  combination of the two  modalities, it is
clearly outperformed  by our descriptor,  as taking the single  nearest neighbor
exhibits a pose error  below $20^\circ$ for $96.2\%$ of the  test samples and an
overall recognition rate of $99.8\%$, an almost perfect score.

\subsection{Generalization}

\begin{figure}[tbp]
\center
\includegraphics[height=0.50\columnwidth,keepaspectratio=true]{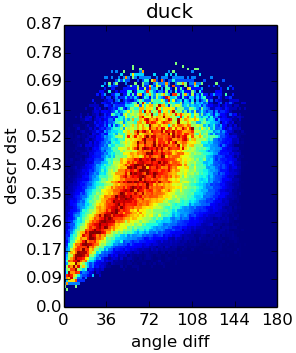} \, %
\includegraphics[height=0.50\columnwidth,keepaspectratio=true]{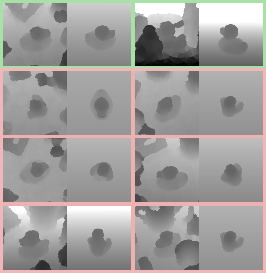}%
\caption{Generalization  to   objects  not seen  during  training.    {\bf  Left:}
  Histogram of correlation  between Pose Similarity and  Descriptor Distance for
  the \emph{duck} that was not included during training for this experiment. The
  recognition rate  is only slightly reduced  compared to when the  duck is used
  during training.  {\bf  Right:} Difficult examples of correct and mis-classifications.   The duck is mostly confused with the ape, which looks  similar in the depth image under some angles.}
\label{fig:o14_plusduck}%
\end{figure}

As a last experiment, we show  that  our  descriptor can  generalize  to  unseen objects.   This evaluation was performed using depth only.  To do so, we train the CNN on 14 out
of the 15  objects.  We then perform  the evaluation just as  above by computing
descriptors  for  the  new  object.   As  can be  seen  from  the  histogram  of
Fig.~\ref{fig:o14_plusduck}-left,  our method  generalizes well  to this  unseen
object.   The overall  performance rate  is slightly  reduced since  the network
could not learn the  subtle differences between the unseen object
and the others.  Most of the miss-classifications are with  the ape, whose shape
looks   similar   to  the   duck's   under   some   viewpoints,  as   shown   in
Fig.~\ref{fig:o14_plusduck}-right.   

\section{Conclusion}

We have shown how to train a CNN  to map raw input images from different input modalities to very compact output descriptors using pair-wise and triplet-wise  constraints over training data and template views.  Our descriptors significantly outperform LineMOD and HOG, which are widely used for object recognition and  3D pose estimation, both in terms of accuracy and descriptor length. 
Our  representation therefore replaces them advantageously. 
Tests on the capability to generalize to unseen objects also have shown promising results.
For further investigation we will make our code available upon request.

{\small
\bibliographystyle{ieee}
\bibliography{string,vision,pwohlhart}
}

\end{document}


\title{Supplementary Material for the paper \\ Learning Descriptors for Object Recognition and 3D Pose Estimation }

\author{Paul Wohlhart and Vincent Lepetit\\
Institute for Computer Vision and Graphics, 
Graz University of Technology, Austria\\
{\tt\small \{wohlhart,lepetit\}@icg.tugraz.at}
}

\maketitle
\thispagestyle{empty}

\section{Additional Samples}

\begin{figure}[h]
\center
\includegraphics[width=0.7\textwidth]{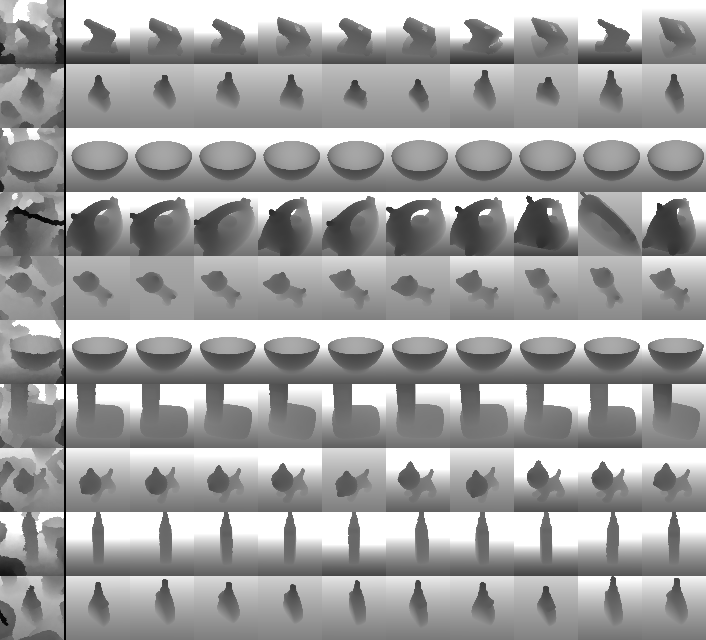} 
\caption{Templates with closest descriptors for samples. Network trained on depth data.}
\label{fig_samples_dpt}
\end{figure}

\begin{figure}[h]
\center
\includegraphics[width=0.7\textwidth]{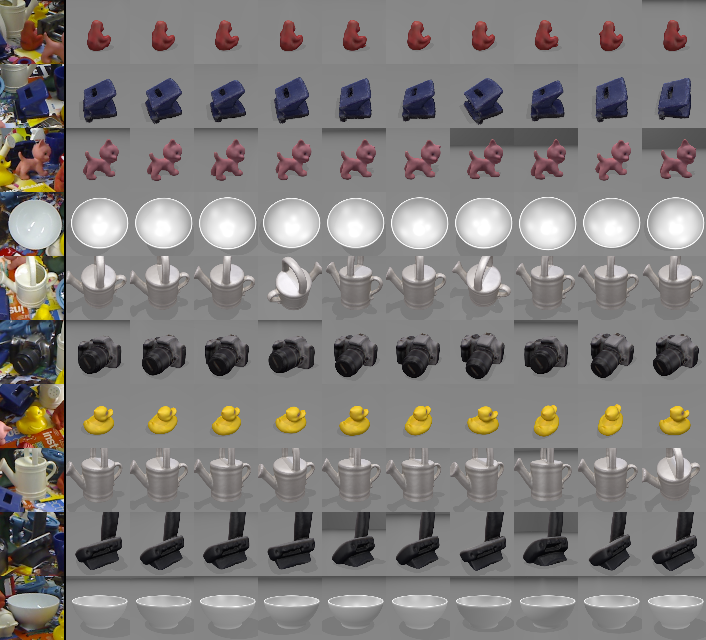} 
\caption{Templates with closest descriptors for samples. Network trained on RGB color data.}
\label{fig_samples_rgb}
\end{figure}

\begin{figure}[h]
\center
\includegraphics[width=0.9\textwidth]{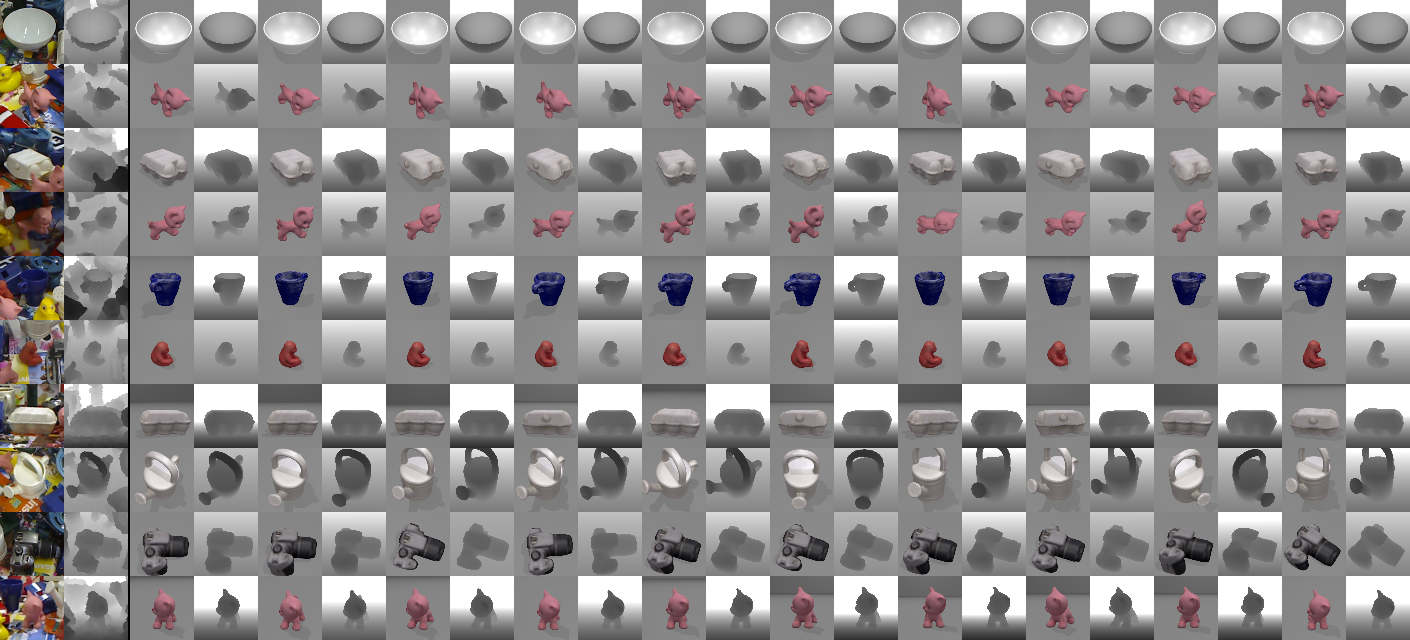} 
\caption{Templates with closest descriptors for samples. Network trained on RGBD data.}
\label{fig_samples_rgbd}
\end{figure}

Figures~\ref{fig_samples_dpt}, \ref{fig_samples_rgb} and \ref{fig_samples_rgbd} show additional examples of templates retrieved for random sets of test samples.
The first column shows the test sample. To the right, each row shows the first 10 templates, sorted by descriptor distance. Note how most of the closest templates show very similar views of the correct object that all give a good estimate of the object's pose.

\clearpage
\newpage
\newpage
\pagebreak
\section{Triplet Cost}

\begin{figure}[h]
\center
\subfloat[our definition]{
\includegraphics[width=0.51\textwidth]{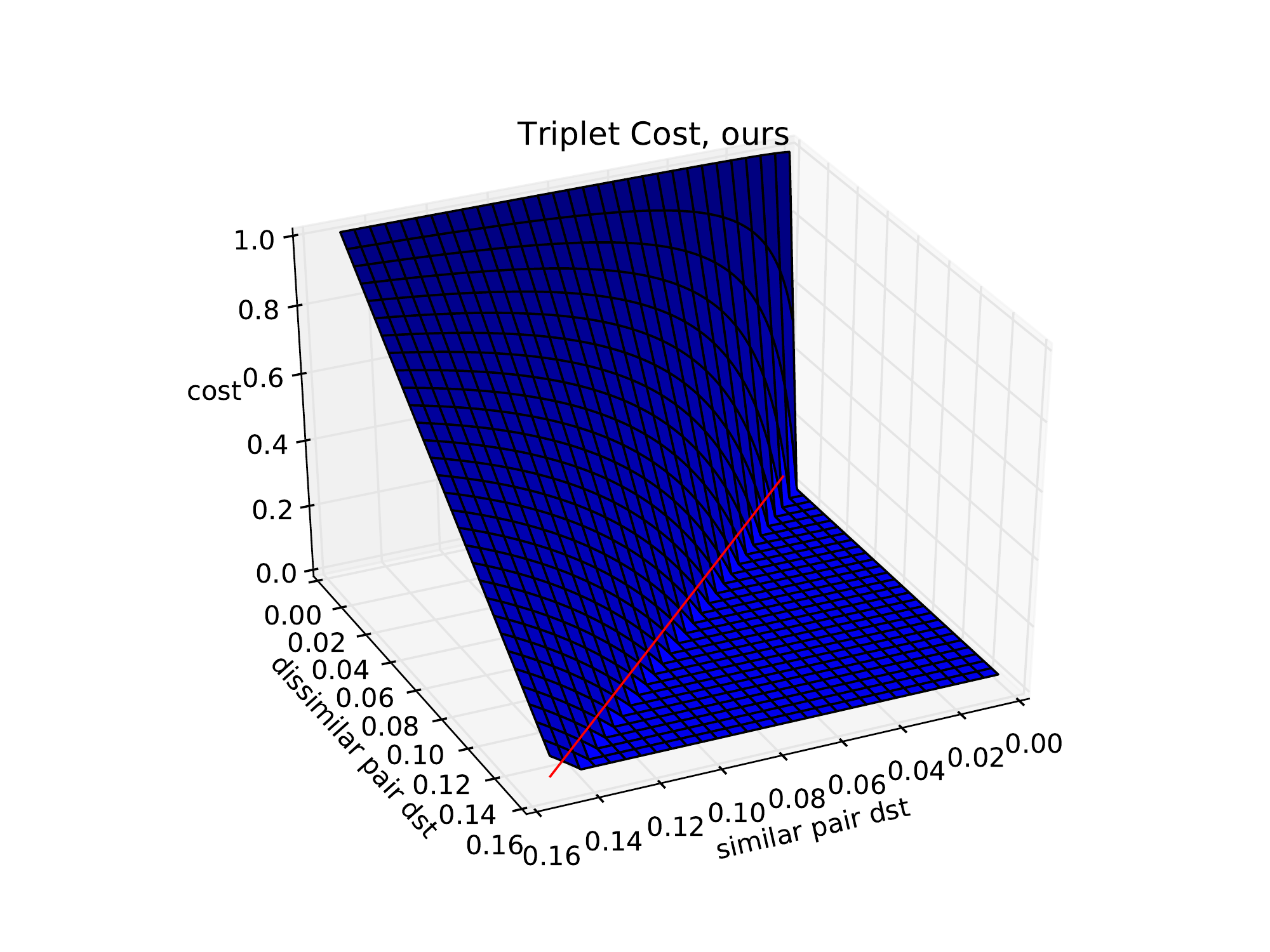} } \\
\subfloat[ours with squared distances]{
\includegraphics[width=0.49\textwidth]{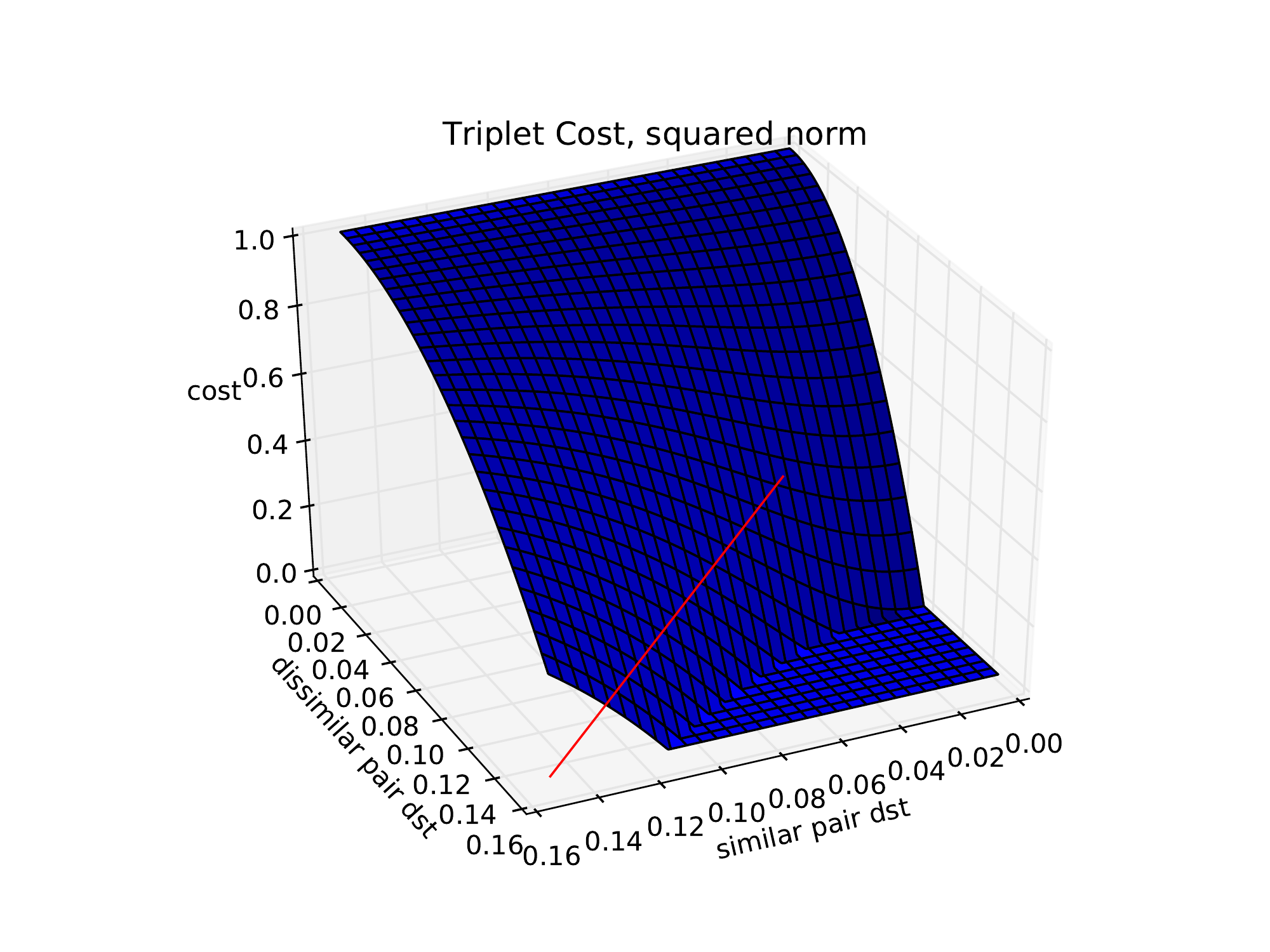} }
\subfloat[Wang et al. CVPR'14]{
\includegraphics[width=0.49\textwidth]{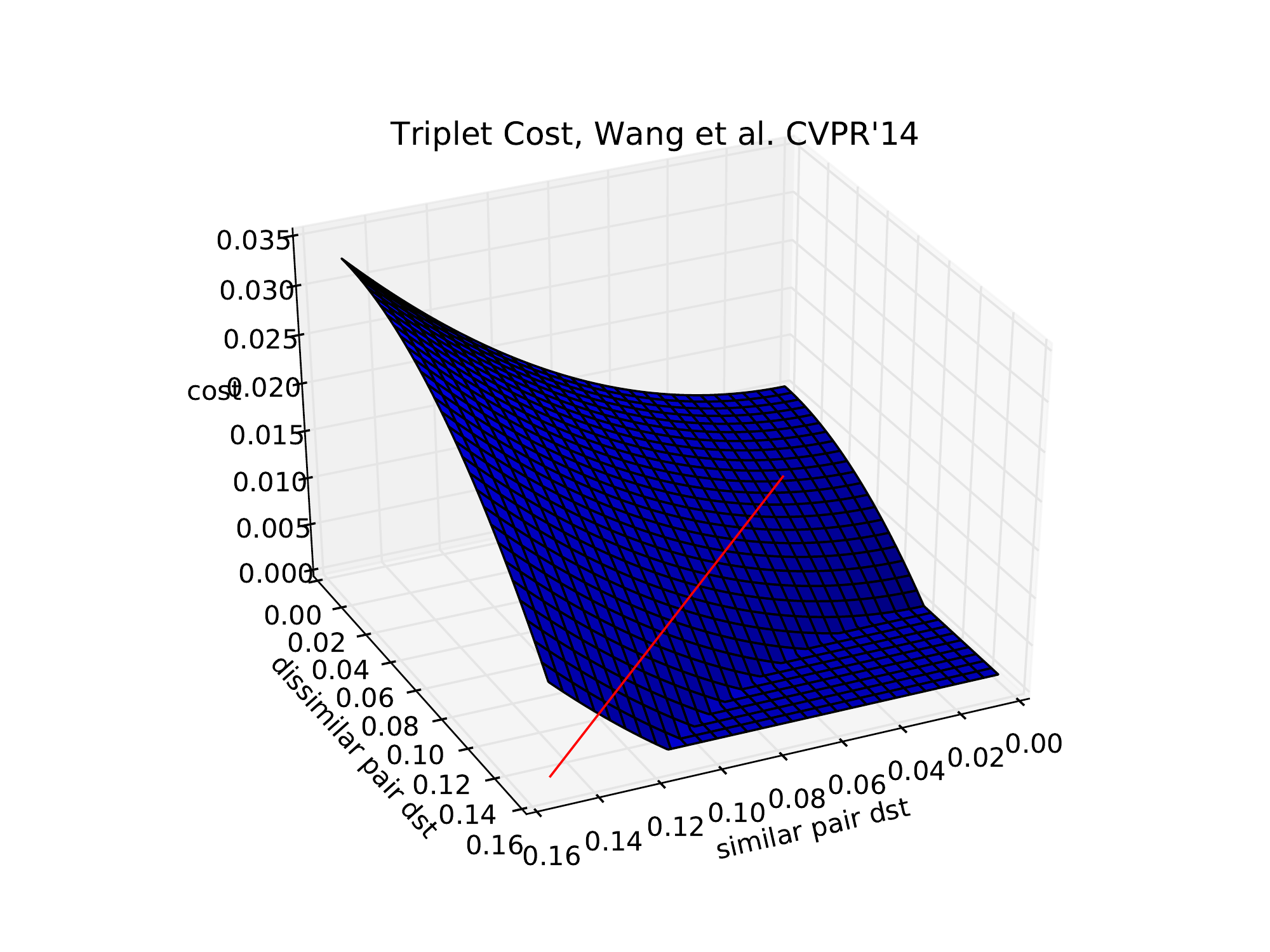} }
\caption{Cost of a triplet for different definitions.}
\label{fig_triplet_cost}
\end{figure}

In Sections 3.2.1 and 3.3 we discuss our definition and implementation of the cost of a triplet in contrast to the definition in related work. Figure~\ref{fig_triplet_cost} shows the value of the cost of one triplet given the distances between similar and dissimilar samples on the x- and y-axis, respectively. On top is our definition, on the lower left our definition, but with the distance squared and on the lower right the definition of Wang \etal in CVPR'14.
As can be seen, in our definition the value of the cost does not depend on the total scale of the triplet.
This allows us to define triplets over arbitrary ranges.
A triplet reaching across the whole template dome does not dominate small local triplets and does not contract the similar pair more than it pushes apart the dissimilar one.

Additionally, like our definition, the other two versions correctly assign high cost to triplets that have a very low distance between the descriptors of the dissimilar samples. However, since the square of the distances is taken, when the distance of the dissimilar pair approaches zero, the derivative w.r.t the distance of the dissimilar pair goes to zero, thus, not pushing apart dissimilar pairs when they are violating the constraints the most.

\newcommand{\Fcnn}{f_{\!_w\!}}
\newcommand{\Lpairs}{\mathcal{L}_\text{pairs}}
\newcommand{\Ltriplets}{\mathcal{L}_\text{triplets}}

